\documentclass[11pt]{article}

\usepackage[preprint]{acl}

\usepackage{times}
\usepackage{latexsym}

\usepackage[T1]{fontenc}

\usepackage[utf8]{inputenc}

\usepackage{microtype}

\usepackage{inconsolata}
\usepackage{booktabs}

\usepackage{graphicx}

\usepackage{svg}

\title{Transformer Enhanced Relation Classification: A Comparative Analysis of Contextuality, Data Efficiency and Sequence Complexity}

\author{Bowen Jing \\
  University of Manchester\\
   \\\And
  Yang Cui\\
  University of Manchester\\
   \\\And
  Tianpeng Huang\\
  University of Manchester\\
  }

\begin{document}
\maketitle

\begin{abstract}

In the era of large language model, relation extraction (RE) plays an important role in information extraction through the transformation of unstructured raw text into structured data \cite{wadhwa2023revisiting}. In this paper, we systematically compare the performance of deep supervised learning approaches without transformers and those with transformers. We used a series of non-transformer architectures such as PA-LSTM\cite{zhang2017position}, C-GCN\cite{zhang2018graph}, and AGGCN(attention guide GCN)\cite{guo2019attention}, and a series of transformer architectures such as BERT, RoBERTa, and R-BERT\cite{wu2019enriching}. Our comparison included traditional metrics like micro F1, as well as evaluations in different scenarios, varying sentence lengths, and different percentages of the dataset for training. Our experiments were conducted on TACRED, TACREV, and RE-TACRED. The results show that transformer-based models outperform non-transformer models, achieving micro F1 scores of 80-90\% compared to 64-67\% for non-transformer models. Additionally, we briefly review the research journey in supervised relation classification and discuss the role and current status of large language models (LLMs) in relation extraction.
\end{abstract}

\section{Introduction}
Relation extraction (RE) \cite{hendrickx2019semeval}, a crucial component of information extraction (IE) \cite{cardie1997empirical}, has gained significant attention for its extensive applications in question answering \cite{xu2016question}, biomedical knowledge discovery \cite{quirk2016distant}, and knowledge base population \cite{zhang2017position}. The RE process generally encompasses two stages: relation identification (RI) and relation classification (RC) \cite{bassignana2022you}. As a vital task within RE, RC aims to identify relationships between pairs of entities in textual data, thereby converting unstructured facts into structured triplet formats (entity1, relation, entity2). For instance, given the sentence “Obama was born in Honolulu” and the entity pair (“Obama”, “Honolulu”), an RE model can identify the relation LOC: city of birth. Traditional methods for RC include rule-based methods, weakly supervised methods, supervised methods, and unsupervised methods.

Recently, using LLMs for relation classification has become popular. LLMs utilize in-context learning due to their flexibility and simplicity\cite{wadhwa2023revisiting}. However, despite these advantages, LLMs often do not perform as well as fully supervised baselines (e.g., BERT or RoBERTa). Additionally, LLMs face challenges related to high computational cost\cite{wadhwa2023revisiting} and data privacy\cite{yao2024survey} concerns. Therefore, exploring alternative methods to LLMs is necessary. Notably, systematic comparative evaluations between deep supervised approaches are lacking. This study aims to fill that gap through an empirical comparison\cite{xu2020brief}.

\begin{figure}
\centering
\includegraphics[width=0.5\textwidth]{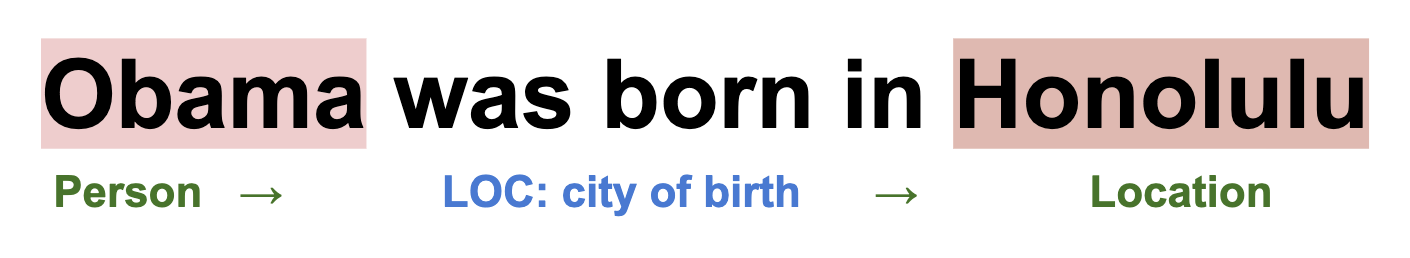}
\label{fig:Tacred Dataset Example}
\caption{Example of Relation Extraction Dataset}
\end{figure}

We specifically concentrate on RC targeting the task of supervised extraction of intra-sentence binary relations in English news corpora. 

In our empirical study, we evaluate non-transformer models and transformer models by answering the following questions:
\begin{itemize}
    \item How much data does each model require to generalize effectively?
    \item How does sentence length affect the model's performance?
\end{itemize}

\section{Related Work}

\paragraph{Deep Learning Models}
Unlike rule-based and SVM methods, deep learning approaches for relationship extraction can automatically learn complex features from data, offering greater adaptability\cite{kumar2017survey} and reducing the need for manual feature engineering and pre-defined domain-specific knowledge\cite{pawar2017relation}. CNNs mark a significant shift towards learning hierarchical representations of text data. Pioneering work, such as that by Zeng et al., demonstrated the advantages of CNNs in capturing lexical and sentence-level features associated with entities within sentences, thereby reducing the necessity for extensive feature engineering.\cite{zeng2014relation}. Following these early efforts, there has been an emergence of studies focusing on CNN derivatives, such as the Multi-window CNN\cite{nguyen2015relation} and the CNNs that performs classification by ranking (CR-CNN) as introduced by Santos et al.\cite{santos2015classifying}.

Sequence models(such as LSTM) gains attention following the success of CNNs. Zhang et al.\cite{zhang2015relation} highlighted the shortcomings of CNNs---- point out CNN models lack the ability to learn temporal features and their difficulties in capturing long-term dependencies. Zhou et al. advanced this field by integrating attention mechanisms within bidirectional LSTM\cite{zhou2016attention}. Building on the success of sequence models, the Position-aware LSTM model introduced by Zhang et al. integrates position information into LSTM\cite{zhang2017position}.

Inspired by sequence models, Graph Convolutional Networks (GCNs) revolutionize RE by efficiently encoding dependency-parsed sentences and enabling parallel computation. Originally implemented in the RE domain by Zhang et al.\cite{zhang2018graph}, the innovative C-GCN model combines a bidirectional LSTM for sequence contextualization with a GCN layer for dependency structure encoding. This model employs a distinctive pruning strategy, enhancing its effectiveness. Following this, Zhou et al. introduced an enhanced C-GCN model utilizing a distance-weighted adjacency matrix, which more accurately represents the dependency tree's structure, and integrates entity attention along with network depth to improve analytical capabilities\cite{zhou2020weighted}. Meanwhile, despite challenges in concentrating on SDP's critical nodes, Guo et al. have advanced the field by pioneering the use of dynamic matrices. This approach enables a more adaptable analysis of the dependency tree through a multi-head self-attention mechanism\cite{guo2019attention}.

\paragraph{BERT Based Language Models}
The introduction of the transformer model by Vaswani et al. in 2017 marked a revolutionary shift in the domain of natural language processing (NLP)\cite{vaswani2017attention}. Its self-attention mechanism not only addresses the challenge of long-range dependencies inherent in sequence-based models such as RNNs and LSTMs but also boasts lower computational complexity for sequences shorter than the embedding size. Crucially, the computation of the attention layers can be performed in parallel, making the training of large language models more feasible and efficient.

Devlin et al. introduced the Bidirectional Encoder Representations from Transformers (BERT)\cite{devlin2018bert}, a new language model pre-trained using a bidirectional masked language model and next sentence prediction tasks on a large corpus. This pre-training process significantly enhances BERT's ability to understand context. Soares et al. demonstrated that fine-tuning BERT leads to substantial improvements in performance over previous models in tasks such as relation extraction, highlighting the model's advanced understanding of language nuances\cite{soares2019matching}.

Building on BERT's success, Wu et al. used an enhanced input representation by substituting the entity mask with an entity marker to emphasize the entities in the sentence that need to be predicted, achieving better results. Later, Zhou et al. added entity type information along with the entity marker and conducted experiments on both BERT and RoBERTa, achieving state-of-the-art results. Furthermore, Meta trained SpanBERT to address relation extraction problems. Unlike BERT, which uses self-supervised learning by masking one word and predicting it, SpanBERT masks a span and predicts the entire span.

\paragraph{GPT based Language Models}
LLMs are undoubtedly the most important topic in the field today, demonstrating significant potential in many areas. However, despite their versatility with in-context learning, LLMs still underperform compared to supervised learning with pre-trained language models (e.g., BERT-based models). Nonetheless, Wan et al. proposed GPT-RE, which nearly achieves SOTA performance. Their method overcomes the low relevance between entities and relations caused by in-context learning by incorporating task-aware representations\cite{wan2023gpt}. This demonstrates that LLMs still have significant potential in RE that remains to be fully explored.

\section{Methodology}

\subsection{Experiment Setting}
We have conducted several experiments to compare the distinct capabilities of each model in relation extraction on the TACRED, TACREV and Re-TACRED datasets. Except the traditional metric(precision, recall, micro F1), we also evaluate these models on test datasets of varying sentence lengths\cite{zhang2017position}. This approach allows us to assess the models' performance across different text lengths. Finally, we train the models on 20\%, 40\%, 60\%, and 80\% of the training dataset. This is to measure the models' performance with limited data, which may indicate how well each model adapts to different downstream tasks\cite{brody2021towards}. Before conducting the experiments, we perform hyperparameter optimization on all experimental models using the grid search method to identify the optimal combination of hyperparameters.

\subsection{Data Process}

\paragraph{Deep learning without transformer models}
In the prepossessing pipeline, we first normalize the raw tokens by converting them to lowercase to ensure uniformity. Next, we replace the entities names with their entity types, such as 'SUBJ-ORG' for orgnazitions.

For integrating the GloVe embeddings\cite{pennington2014glove}, we normalize tokens to align with GloVe's format, for example, converting '(' to '-LRB-‘. we construct our vocabulary by combining tokens from our dataset with those found in GloVe, applying a frequency threshold to filter out rare words. This amalgamated vocabulary includes both special tokens and entity mask tokens, the latter derived from the named entity recognition (NER) tags provided by the TACRED dataset, such as 'SUBJ-PERSON' for personal entities. In the end, we construct word embeddings by mapping out vocabulary to pre-trained Glove vectors, initialzing with random value where no match is found.

\paragraph{Deep learning with transformer models}
In our data processing strategy, we use a refined version of the typed entity marker technique. For RBERT, we utilize an entity marker approach that designates the subject and object entities with ``@'' and ``\#'' respectively. This method clearly distinguishes the entities within the text without adding any new special tokens. For BERT and RoBERTa, we further enhance this technique by including type information. This involves enclosing the subject and object entities with ``@'' and ``\#'' while also indicating their types with label text. Specifically, the subject type is prefixed with ``*'', and the object type is prefixed with ``\^{}''. The resulting text format is ``@ *subj-type* SUBJECT @ ... \# \^{}obj-type\^{} OBJECT \#'', where ``subj-type'' and ``obj-type'' represent the named entity recognition (NER) labels. This enriched representation aims to enhance the models' ability to understand and extract relationships by explicitly incorporating type information along with entity markers.

\section{Result Analysis}

\begin{table}[htbp]
\centering
\resizebox{0.5\textwidth}{!}{%
\begin{tabular}{l c c|c}
  \hline
  \textbf{Deep Learning Models}  & TACRED & TACREV & Re-TACRED \\
  \hline
  PALSTM & 64.64 & 73.74 &78.63 \\
  C-GCN & 66.15 & 73.99 & 80.62 \\
  Att-Guide-GCN & \textbf{68.97} & \textbf{76.77} & \textbf{80.68} \\

  \hline
  \textbf{Pretrain Language Models} & TACRED & TACREV & Re-TACRED\\
  \hline
  BERT    &72.74&81.39&89.42\\
  RoBERTa &\textbf{74.84}&\textbf{83.09}&\textbf{91.53}\\
  R-BERT &70.09&78.86&89.22\\
  \hline
  \textbf{GPT based Models} & TACRED & TACREV & Re-TACRED\\
  \hline
  GPT-RE\cite{wan2023gpt} &72.14& - & -   \\
  \hline
\end{tabular}%
}
\caption{\label{tab:F1 of all model on all dataset}(micro)F1 score of different deep learning models and language models trained with TACRED and evaluated with TACRED and TACREV datasets. On the right-hand side is the F1 score of models trained with Re-TACRED and also evaluate with it}
\end{table}

The table \ref{tab:F1 of all model on all dataset} presents the results of three deep learning models and three language models. These models were trained on the TACRED dataset and evaluated on both the TACRED and TACREV datasets. Additionally, the models were trained and evaluated on the Re-TACRED dataset. The results indicate that the average F1 scores on Re-TACRED and TACREV are significantly higher than those on TACRED, demonstrating that the two revised TACRED datasets are of higher quality than the original TACRED.

Among the deep learning models, Att-Guide-GCN achieves the highest F1 scores across all three datasets, with a notable score of 80.68 on Re-TACRED. In contrast, PALSTM performs better than C-GCN overall.

For the BERT family models, R-BERT is less competitive than BERT. However, RoBERTa outperforms all other models, achieving the highest F1 scores across the three evaluation datasets, with an impressive score of 91.53 on Re-TACRED.

\begin{figure}[htbp]
\centering
\includegraphics[width=0.5\textwidth]{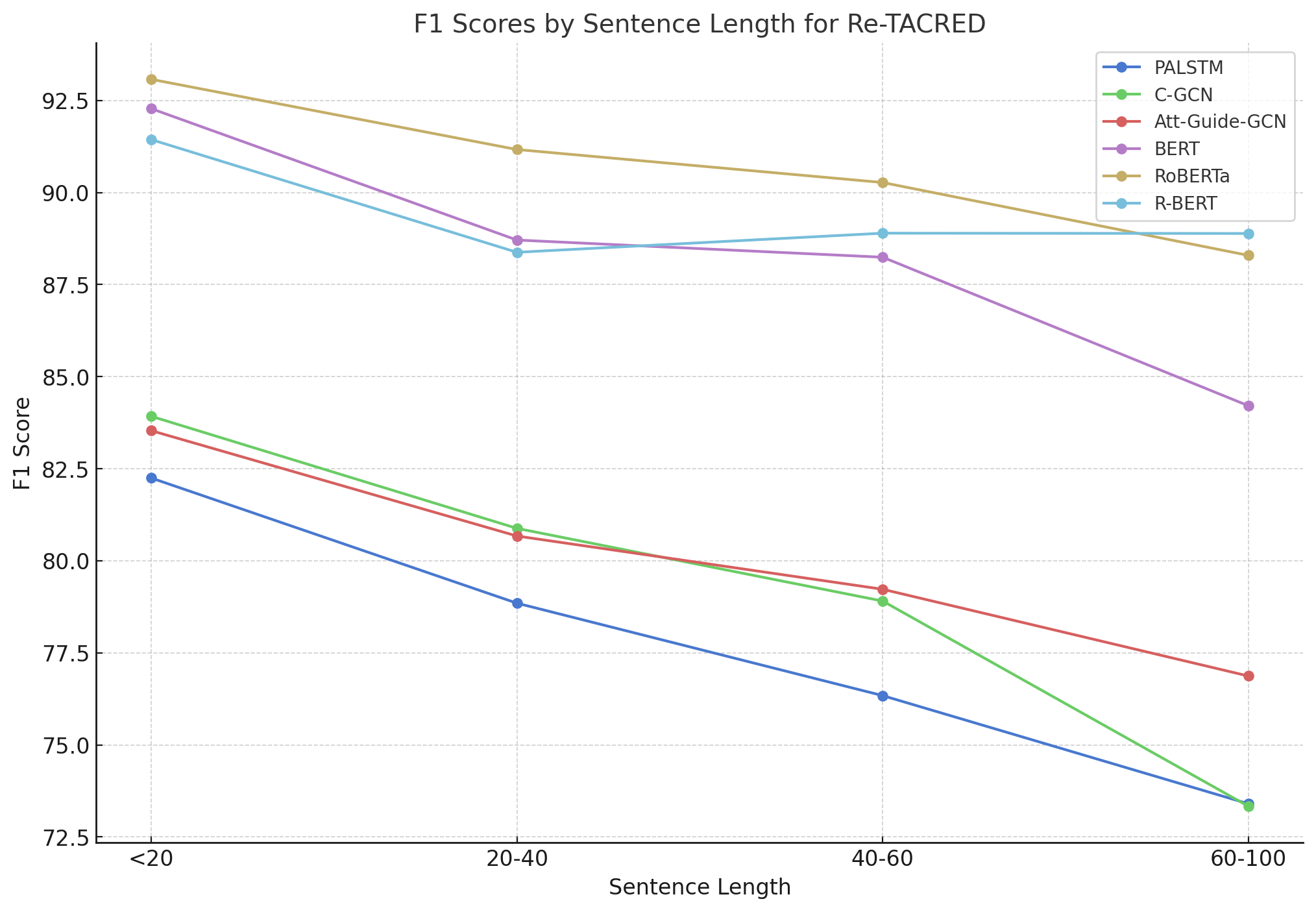}
\caption{Re-TACRED with Different Sentence Lengths}
\label{fig:F1_SentenceLength_Re-TACRED}
\end{figure}

The line chart \ref{fig:F1_SentenceLength_Re-TACRED} illustrates the F1 scores of six models evaluated on Re-TACRED with varying sentence lengths. The results reveal an interesting trend: as sentence length increases from fewer than 20 tokens to 100 tokens, the performance of most models declines. However, R-BERT's performance improves after the sentence length exceeds 40 tokens. Notably, for sentences ranging from 60 to 100 tokens, BERT's performance is almost five points lower than that of the other BERT-based models, indicating that R-BERT and RoBERTa handle longer sequences more effectively than BERT.

Furthermore, there is a significant performance gap between the deep learning models and the BERT-based models. The BERT-based language models consistently outperform the deep learning models across all sentence lengths, demonstrating that language models are more adept at understanding relationships between entities.

\begin{figure}[htbp]
\centering
\includegraphics[width=0.5\textwidth]{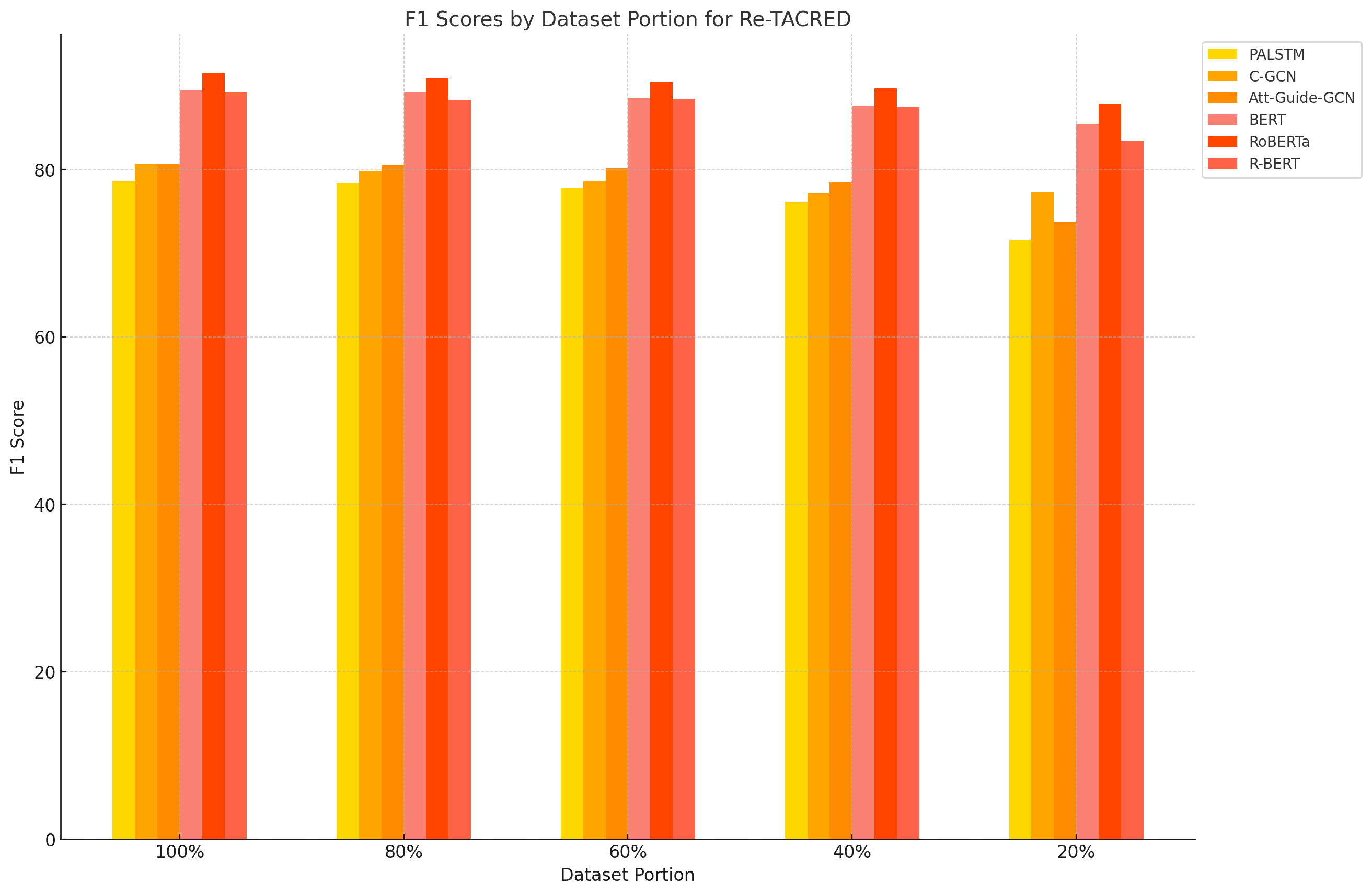}
\caption{Re-TACRED with Different Dataset Portion}
\label{fig:F1_DatasetPortion_TACREV}
\end{figure}

The bar chart \ref{fig:F1_DatasetPortion_TACREV} depicts the third experiment, which evaluates model performance using different portions of training data by measuring the F1 score. This experiment aims to assess model performance in few-shot learning scenarios.

The chart shows that as the portion of the dataset increases, the performance of all six models improves. Notably, there is a significant performance boost between the 20\% and 40\% training sets for the BERT-based models, indicating that these models learn faster even with limited data. Additionally, the performance gap between the deep learning models and the BERT-based models is more pronounced when less training data is available. This suggests that BERT-based language models are more efficient in understanding relationships between entities with fewer training samples.

\section{Conclusion and Future Work}
The above results demonstrate that BERT-based models consistently outperform other deep learning models in relation extraction tasks under various conditions. Specifically, RoBERTa achieves an impressive F1 score of 91.53 on the Re-TACRED dataset, a performance level that matches or exceeds human capability. These models also require fewer computational resources for both training and inference compared to larger, more general language models. Additionally, the ability to run BERT-based models on local machines offers significant advantages in terms of data privacy, which is crucial for handling sensitive information such as clinical data.

However, despite the growing popularity of modern language models like GPT-3, they exhibit notable limitations. The GPT-based model GPT-RE does not perform as well as supervised baseline models in our experiments, as detailed in Table \ref{tab:F1 of all model on all dataset}. Furthermore, Wadhwa et al. have highlighted the high cost associated with prompting GPT-3, making it an expensive choice for large-scale or frequent querying tasks\cite{wadhwa2023revisiting}. These findings underscore the need for careful consideration when selecting models for relation extraction tasks, balancing performance, computational cost, and data privacy requirements.

\bibliography{custom}
\appendix

\section{Appendix}
\label{sec:appendix}
\subsection{Dataset Comparison}
\begin{table}[h!]
\centering
\begin{tabular}{lrrrrr}
\toprule
\hline
\hline
Dataset & \# train & \# dev & \# test & \# classes \\
\midrule
\hline
TACRED    & 68124 & 22631 & 15509 & 42 \\
\hline
TACREV    & - & 22631 & 15509 & 42 \\
\hline
Re-TACRED & 58465 & 19584 & 13418 & 40 \\
\bottomrule
\hline
\hline
\end{tabular}
\caption{Summary of datasets used in the experiments.}
\label{tab:datasets}
\end{table}

\subsection{Experiment Results for TACRED and TACREV}
\begin{figure}[!h]
\centering
\includegraphics[width=0.5\textwidth]{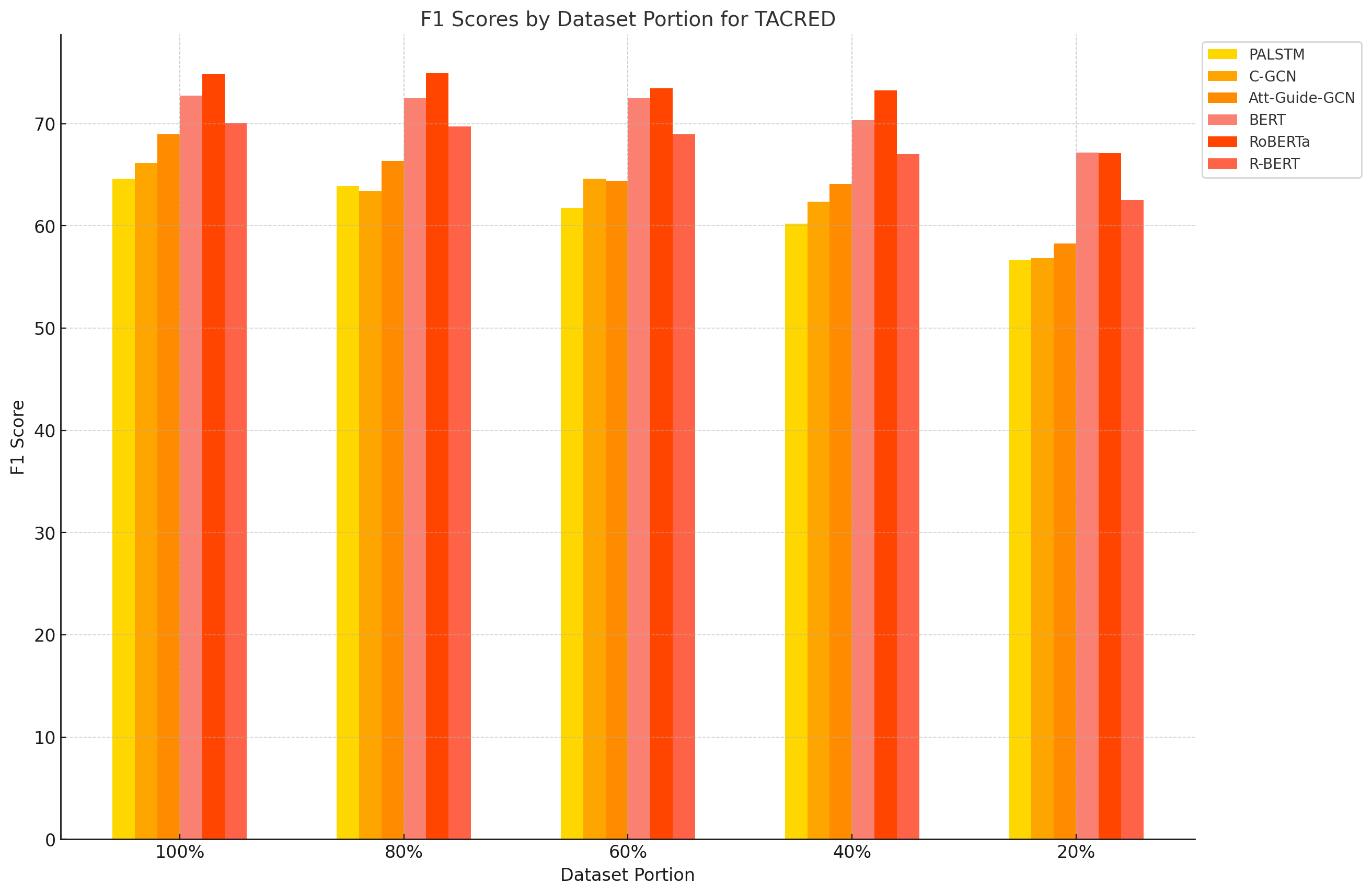}
\caption{TACRED with  Different Dataset Portion}
\label{fig:F1_SentenceLength_Re-TACRED}
\end{figure}

\begin{figure}[!h]
\centering
\includegraphics[width=0.5\textwidth]{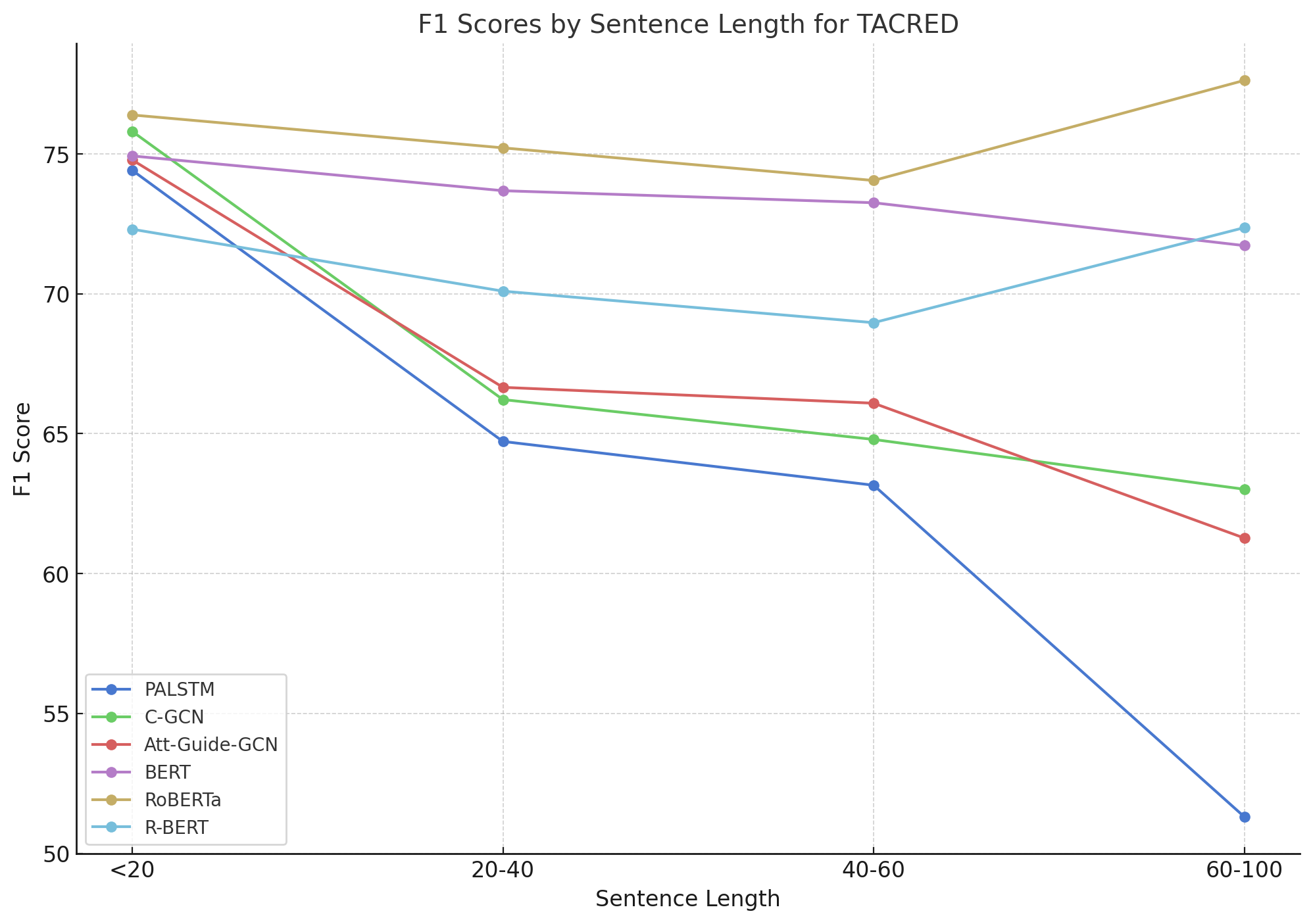}
\caption{TACRED with Different Sentence Lengths}
\label{fig:F1_SentenceLength_Re-TACRED}
\end{figure}

\begin{figure}[!h]
\centering
\includegraphics[width=0.5\textwidth]{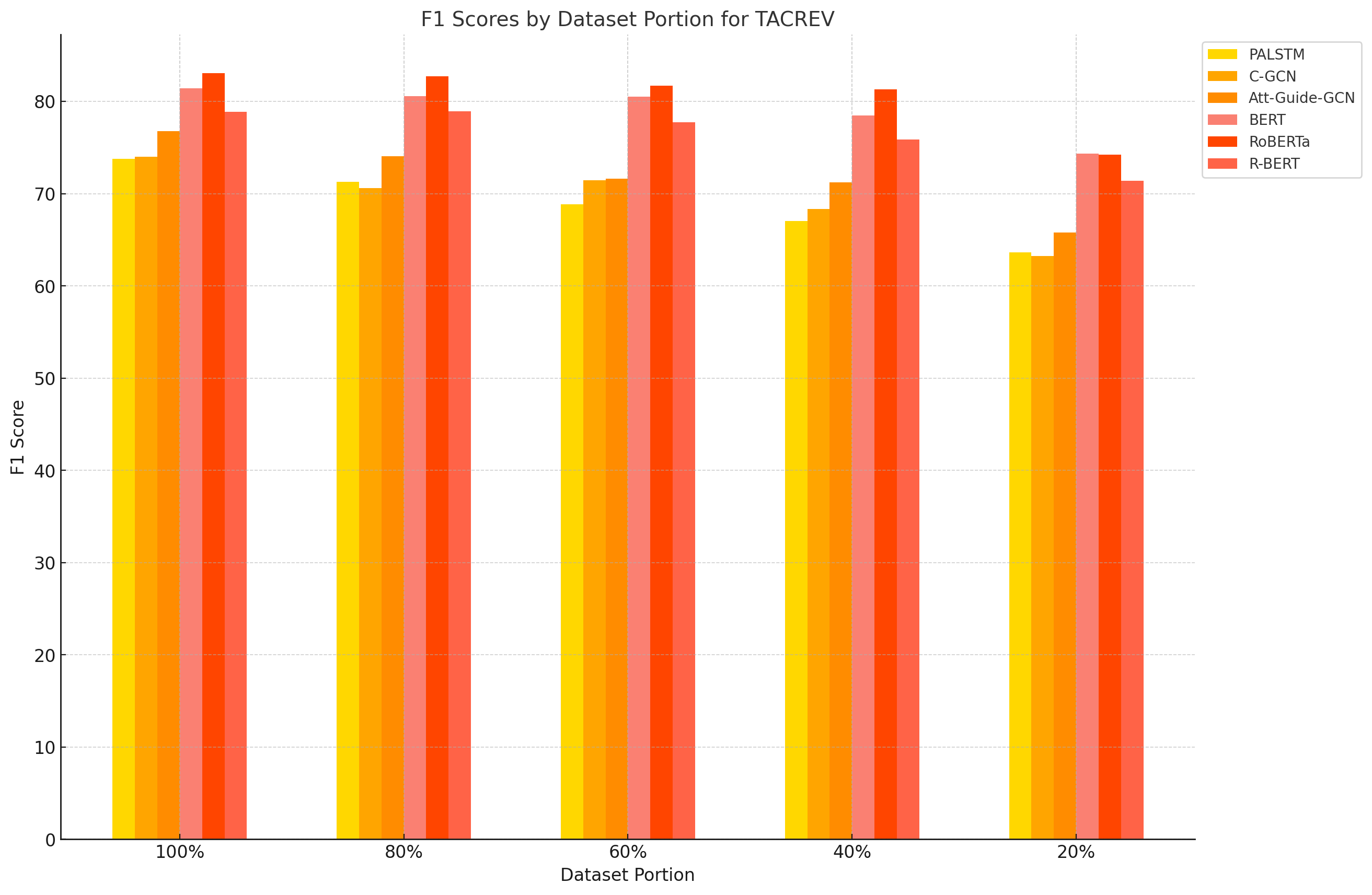}
\caption{TACREV with  Different Dataset Portion}
\label{fig:F1_SentenceLength_Re-TACRED}
\end{figure}

\begin{figure}[!h]
\centering
\includegraphics[width=0.5\textwidth]{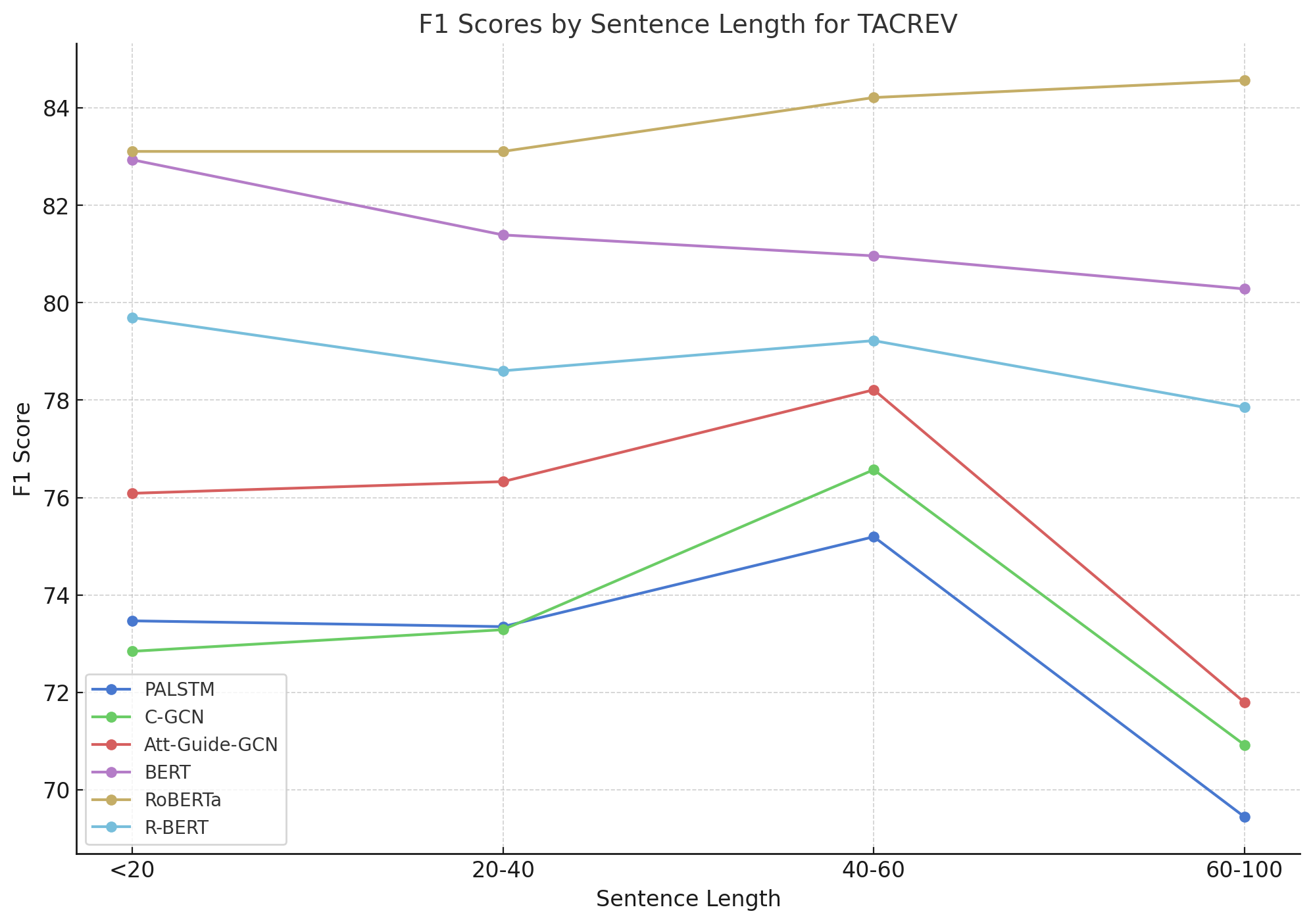}
\caption{TACREV with Different Sentence Lengths}
\label{fig:F1_SentenceLength_Re-TACRED}
\end{figure}
Our experiment is divided into two parts. In the first part, we train on the TACRED training set and validate on both the TACRED validation set and TACREV. Re-TACRED is trained and validated separately. The Re-TACRED dataset is a significant improvement over the TACRED relation extraction dataset. With new crowd-sourced labels, Re-TACRED prunes poorly annotated sentences and addresses the ambiguities in TACRED's relation definitions, ultimately correcting 23.9\% of the TACRED labels.
\end{document}